\documentclass[letterpaper]{article} % DO NOT CHANGE THIS
\usepackage{aaai24}  % DO NOT CHANGE THIS
\usepackage{times}  % DO NOT CHANGE THIS
\usepackage{helvet}  % DO NOT CHANGE THIS
\usepackage{courier}  % DO NOT CHANGE THIS
\usepackage[hyphens]{url}  % DO NOT CHANGE THIS
\usepackage{graphicx} % DO NOT CHANGE THIS
\usepackage{natbib}  % DO NOT CHANGE THIS AND DO NOT ADD ANY OPTIONS TO IT
\usepackage{caption} % DO NOT CHANGE THIS AND DO NOT ADD ANY OPTIONS TO IT
\usepackage{algorithm}
\usepackage{algorithmic}

\usepackage{xcolor}
\usepackage{paralist}
\usepackage{enumitem}
\usepackage{xspace}
\usepackage{colortbl}
\usepackage{color}
\usepackage{amssymb}
\definecolor{citecolor}{HTML}{0b64c5}
\definecolor{cello}{HTML}{ffe6cc}

\usepackage{newfloat}
\usepackage{listings}
\DeclareCaptionStyle{ruled}{labelfont=normalfont,labelsep=colon,strut=off} % DO NOT CHANGE THIS
\floatstyle{ruled}
\newfloat{listing}{tb}{lst}{}
\floatname{listing}{Listing}
\title{BLIVA: A Simple Multimodal LLM for Better Handling of Text-Rich Visual Questions}
\author {
    Wenbo Hu\equalcontrib \textsuperscript{\rm 1},
    Yifan Xu\equalcontrib \textsuperscript{\rm 2},
    Yi Li\textsuperscript{\rm 1}
    Weiyue Li\textsuperscript{\rm 1}
    Zeyuan Chen\textsuperscript{\rm 1}
    Zhuowen Tu\textsuperscript{\rm 1}
}
\affiliations {
    \textsuperscript{\rm 1}UC San Diego \ 
    \textsuperscript{\rm 2}Coinbase Global, Inc.\\
   \{w1hu, yil115, wel019, zec016, ztu\}@ucsd.edu \quad yifan.xu@coinbase.com
}

\usepackage{bibentry}
\newcommand{\modelName}{BLIVA\xspace}
\definecolor{mygray}{gray}{.90}
\newcommand{\colorgray}{\cellcolor{mygray}}
\newcolumntype{g}{>{\columncolor{mygray}}c}

\begin{document}

\maketitle

\begin{abstract}
Vision Language Models (VLMs), which extend Large Language Models (LLM) by incorporating visual understanding capability, have demonstrated significant advancements in addressing open-ended visual question-answering (VQA) tasks. However, these models cannot accurately interpret images infused with text, a common occurrence in real-world scenarios. Standard procedures for extracting information from images often involve learning a fixed set of query embeddings. These embeddings are designed to encapsulate image contexts and are later used as soft prompt inputs in LLMs. Yet, this process is limited to the token count, potentially curtailing the recognition of scenes with text-rich context. To improve upon them, the present study introduces \textbf{\modelName}: an augmented version of Instruct\textbf{BLI}P with \textbf{V}isual \textbf{A}ssistant. \modelName incorporates the query embeddings from InstructBLIP and also directly projects encoded patch embeddings into the LLM, a technique inspired by LLaVA. This approach assists the model to capture intricate details potentially missed during the query decoding process. Empirical evidence demonstrates that our model, \modelName, significantly enhances performance in processing text-rich VQA benchmarks (up to  17.76\% in OCR-VQA benchmark) and in undertaking general {(not particularly text-rich)} VQA benchmarks (up to 7.9\% in Visual Spatial Reasoning benchmark), and achieved 17.72\% overall improvement in a comprehensive multimodal LLM benchmark (MME),  comparing to our baseline InstructBLIP. \modelName demonstrates significant capability in decoding real-world images, irrespective of text presence.  To demonstrate the broad industry applications enabled by \modelName, we evaluate the model using a new dataset comprising YouTube thumbnails paired with question-answer sets across 11 diverse categories. For researchers interested in further exploration, our code and models are freely accessible at \textcolor{blue}{\url{https://github.com/mlpc-ucsd/BLIVA}}.
\end{abstract}

\begin{figure*}[htbp]
\centering
%\framebox[4.0in]{$\;$}
% \includegraphics[width=\textwidth]{images/Compare_arxiv2.pdf}

\includegraphics[width=0.95\textwidth]{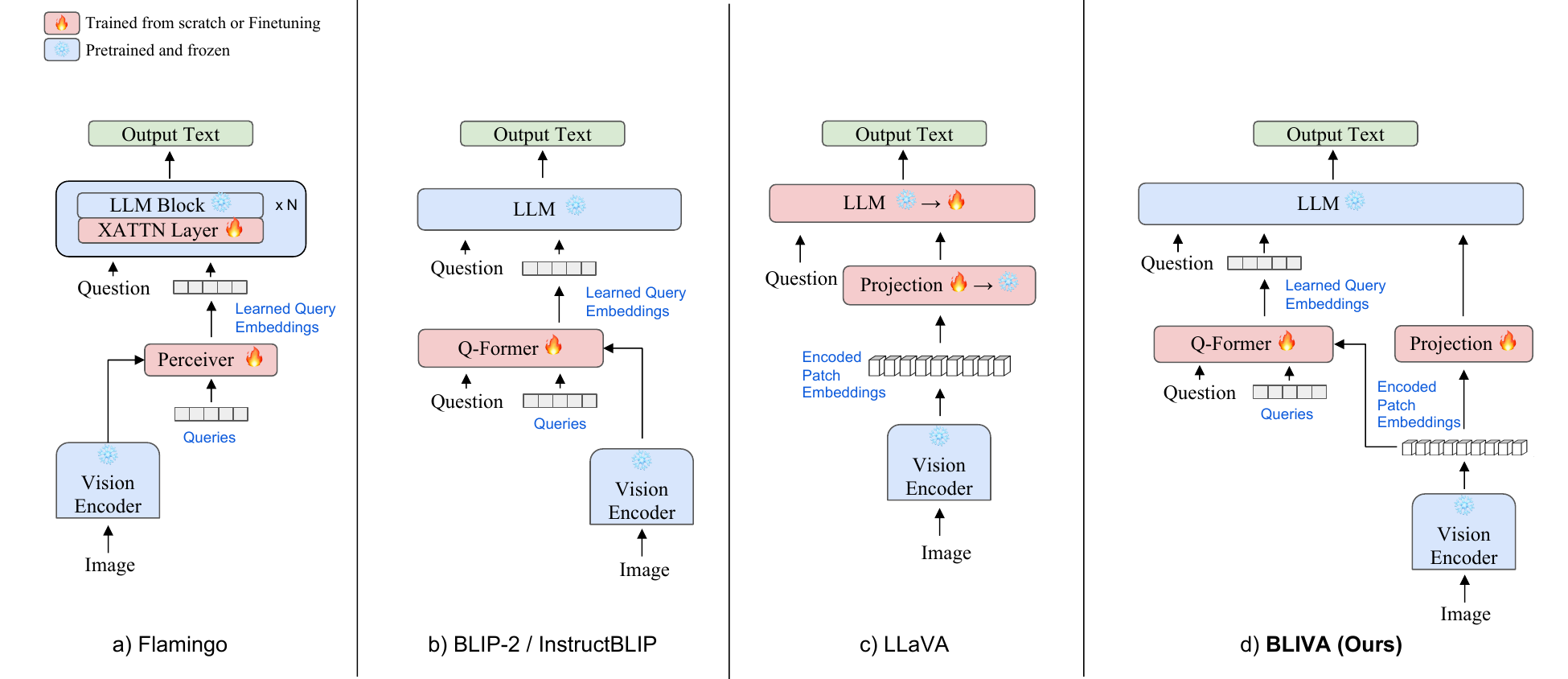}

\caption{\textbf{Comparison of various VLM approaches}. Both (a) Flamingo \citep{Flamingo} and (b) BLIP-2 / InstructBLIP \citep{BLIP-2, insblip} architecture utilize a fixed, small set of query embeddings. These are used to compress visual information for transfer to the LLM. In contrast, (c) LLaVA aligns the encoded patch embeddings directly with the LLM. (d) BLIVA (Ours) builds upon these methods by merging learned query embeddings with additional encoded patch embeddings.}
\label{fig:comparison}
\end{figure*}

\section{Introduction}

 Recently, Large Language Models (LLMs) have transformed the field of natural language understanding, exhibiting impressive capabilities in generalizing across a broad array of tasks, both in zero-shot and few-shot settings. This success is mainly contributed by instruction tuning~\citep{InstructionTuning} which improves generalization to unseen tasks by framing various tasks into instructions. Vision Language Models (VLMs) such as OpenAI's GPT-4~\citep{openai2023gpt4}, which incorporates LLM with visual understanding capability, have demonstrated significant advancements in addressing open-ended visual question-answering (VQA) tasks. Several approaches have been proposed for employing LLMs on vision-related tasks by directly aligning with a visual encoder's patch feature ~\citep{LLaVA} or extracting image information through a fixed number of query embeddings.~\citep{BLIP-2, MiniGPt4}. 

However, despite exhibiting considerable abilities for image-based human-agent interactions, these models struggle with interpreting text within images. Images with text are pervasive in our daily lives, and comprehending such content is essential for human visual perception.  Previous works utilized an abstraction module with queried embeddings, limiting their capabilities in textual details within images~\citep{BLIP-2, openflamingoanas_awadalla_2023_7733589, mPlug-OwL}.

In our work, we employ learned query embeddings with additional visual assistant branches, utilizing encoded patch embeddings. This approach addresses the constraint image information typically provided to language models, leading to improved text-image visual perception and understanding. 
% For enhanced localization ability which requires higher resolution to achieve better performance, we also support scaling the image input resolution from $\displaystyle 224^2 $ to $\displaystyle 384^2 $ by simply interpolating the positional embedding. All the experimental results we currently reported are based on 224 resolution to match with others' work.  %without further training. 
Empirically, we report the results of our model in general {(not particularly text-rich)} VQA benchmarks following the evaluation datasets of \citep{insblip} and text-rich image evaluation protocol from \citep{liu2023hidden}. Our model is initialized from a pre-trained InstructBLIP and an encoded patch projection layer trained from scratch. Following \citep{MiniGPt4, LLaVA}, we further demonstrate a two-stage training paradigm. We begin by pre-training the patch embeddings projection layer. Subsequently, with the instruction tuning data, we fine-tune both the Q-former and the patch embeddings projection layer. During this phase, we maintain both the image encoder and LLM in a frozen state. We adopt this approach based on two findings from our experiments: firstly, unfreezing the vision encoder results in catastrophic forgetting of prior knowledge; secondly, training the LLM concurrently didn't bring improvement but brought significant training complexity. 
%For better localization ability, especially for OCR tasks, positional encoding is critical for 

 In summary, our study consists of the following highlights:
% In summary, our study makes the following contributions:
\begin{itemize}
    \item We present \textbf{\modelName}, which leverages both learned query embeddings and encoded patch embeddings, providing an effective method for interpreting text within images.
%providing a more robust VQA solution for interpreting text within images. (lower the tone)
    \item Our experimental results showcase that \textbf{\modelName} provides improvements in the understanding of text within images while maintaining a robust performance in general {(not particularly text-rich)} VQA benchmarks and achieving the best performance on MME benchmark among previous methods.
    \item To underscore the real-world applicability of \textbf{\modelName}, we evaluate the model using a new dataset of YouTube thumbnails with associated question-answer pairs.
\end{itemize}

\begin{figure*}[htbp]
\centering
\includegraphics[width=0.8\textwidth] {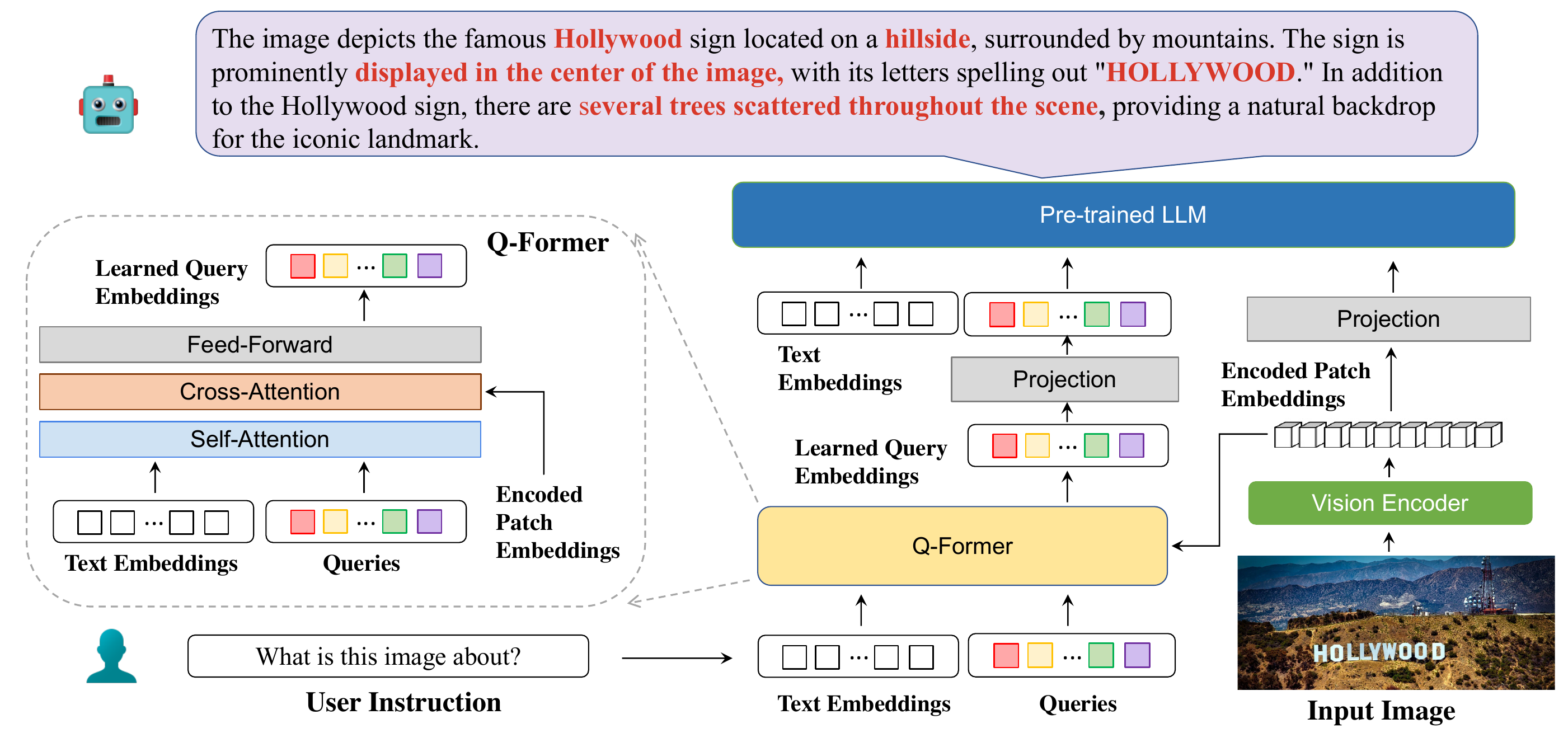}
\caption{ \textbf{Model architecture of BLIVA}. BLIVA uses a Q-Former to draw out instruction-aware visual features from the patch embeddings generated by a frozen image encoder. These learned query embeddings are then fed as soft prompt inputs into the frozen Language-Learning Model (LLM). Additionally, the system repurposes the originally encoded patch embeddings through a fully connected projection layer, serving as a supplementary source of visual information for the frozen LLM.}
\label{fig:detail}
\end{figure*}

\section{Related Work}

\subsection{Multimodal Large Language Model}

Large Language Models (LLMs) have demonstrated impressive zero-shot abilities across various
open-ended tasks. Recent research has explored the application of LLMs for multimodal generation to understand visual inputs. Some approaches leverage the pre-trained LLM to build unified models for multi-modality. For example, Flamingo~\cite{Flamingo} connects the vision encoder and LLM by a Perceiver Resampler which exhibits impressive few-shot performance. Additionally, BLIP-2~\citep{BLIP-2} designs a Q-former to align the visual feature with OPT~\cite{opt} and FLAN-T5~\citep{flanweifinetuned}. MiniGPT-4~\citep{MiniGPt4} employed the same Q-former but changed the LLM to Vicuna~\citep{Vicuna}. Some approaches also finetuned LLM for better alignment with visual features such as LLaVA~\citep{LLaVA} directly finetuned LLM and mPLUG-Owl~\citep{mPlug-OwL} performs low-rank adaption (LoRA)~\citep{hu2022lora} to finetune a LLaMA model~\citep{touvron2023llama}. PandaGPT~\citep{pandagpt} also employed LoRA to finetune a Vicuna model on top of ImageBind~\citep{imagebind}, which can take multimodal inputs besides visual. While sharing the same two-stage training paradigm, we focus on developing an end-to-end multimodal model for both text-rich VQA benchmarks and general VQA benchmarks.

\subsection{Multimodal instruction tuning}

Instruction tuning has been shown to improve the generalization performance of language models
to unseen tasks. In the natural language processing (NLP) community, some approaches collect instruction-tuning data by converting existing NLP datasets into instruction format~\citep{wang2022supernaturalinstructions, flanweifinetuned, sanh2022multitask,chung2022scaling} others use LLMs to generate instruction data~\citep{alpaca, Vicuna, wang2023selfinstruct, honovich2022unnatural}. 
% Instruction data enabled LLMs such as GPT-3~\citep{gpt3brown2020language}, T5~\citep{2020t5} to follow natural language instructions and align with human values, leading to instruction-tuned counterparts such as Instruct GPT~\citep{ouyang2022training}/ChatGPT~\citep{chatgpt} and FLAN-T5~\citep{flanweifinetuned}. 
Recent research expanded instruction tuning to multimodal settings. 
% including audio~\citep{huang2023audiogpt,zhang2023speechgpt}, image, and video~\citep{zhang2023videollama,maaz2023videochatgpt}. 
In particular, for image-based instruction tuning, MiniGPT-4~\citep{MiniGPt4} employs human-curated instruction data during the finetuning stage. LLaVA~\citep{LLaVA} generates 156K multimodal instruction-following data by prompting GPT-4~\citep{openai2023gpt4} with image captions and bounding boxes coordinates. mPLUG-Owl~\citep{mPlug-OwL} also employs 400K mixed text only and multimodal instruction data for finetuning. Instruction tuning also enhanced the previous vision language foundation model's performance. For example, MultimodalGPT~\citep{gong2023multimodalgpt} designed various instruction templates that incorporate vision and language data for multi-modality instruction tuning OpenFlamingo~\citep{openflamingoanas_awadalla_2023_7733589}. \citep{xu2023multiinstruct} built a multimodal instruction tuning benchmark dataset that consists of 62 diverse multimodal tasks in a unified seq-to-seq format and finetuned OFA~\citep{wang2022ofa}. MIMIC-IT~\citep{li2023mimicit} built a bigger dataset comprising 2.8 million multimodal instruction-response pairs to train a stronger model Otter~\citep{li2023mimicit}. We also employed instruction tuning data following the same prompt as InstructBLIP\citep{insblip} to demonstrate the effectiveness of utilizing additional encoded patch embeddings.

\section{Method}

\subsection{Architecture Overview}

As illustrated in Figure \ref{fig:comparison}, there are mainly two types of end-to-end multimodal LLMs: 1) Models that utilize learned query embeddings for LLM. For instance, MiniGPT-4~\citep{MiniGPt4} used the frozen Q-former module from BLIP-2~\citep{BLIP-2} to extract image features by querying the CLIP vision encoder. Flamingo~\citep{Flamingo}, employed a Perceiver Resampler, which reduced image features to a fixed number of visual outputs for LLM. 2) Models that directly employed image-encoded patch embeddings, such as LLaVA~\citep{LLaVA}, which connect its vision encoder to the LLM using an MLP. Nevertheless, these models exhibit certain constraints. Some models employ learned query embeddings for LLM, which help in better understanding the vision encoder but may miss crucial information from encoded patch embeddings. On the other hand, some models directly use encoded image patch embeddings through a linear projection layer, which might have limited capability in capturing all the information required for LLM. 

To address this, we introduce \modelName, a multimodal LLM designed to incorporate both learned query embeddings — which are more closely aligned with the LLM — and image-encoded patch embeddings that carry richer image information.  In particular, Figure \ref{fig:detail} illustrates that our model incorporates a vision tower, which encodes visual representations from the input image into encoded patch embeddings. Subsequently, it is sent separately to the Q-former to extract refined learned query embeddings, and to the projection layer, allowing the LLM to grasp the rich visual knowledge. We concatenate the two types of embeddings and feed them directly to the LLM. These combined visual embeddings are appended immediately after the question text embedding to serve as the final input to the LLM. During inference, we employed beam search to select the best-generated output. Conversely, for classification and multi-choice VQA benchmarks, we adopted the vocabulary ranking method as outlined in InstructBLIP~\citep{insblip}. Given our prior knowledge of a list of candidates, we calculated the log-likelihood for each and chose the one with the highest value as the final prediction. To support another version for commercial usage of our architecture, we also selected FlanT5 XXL as our LLM. This is named as \modelName (FLanT5\textsubscript{XXL}) in this paper.

% To demonstrate \modelName's ability to focus in images with text and OCR tasks. we trained our model with Vicuna as LLM for a version called \modelName Text-Expert, which utilized the training data of the four OCR task datasets and evaluate on their evaluation split. 

% We propose to use patch embeddings which contain localized details for LLM to fully understand the image and answer human questions correctly. 

\subsection{Two stages Training Scheme}
\begin{figure}[htbp]
%\framebox[4.0in]{$\;$}
\includegraphics[width=\columnwidth]{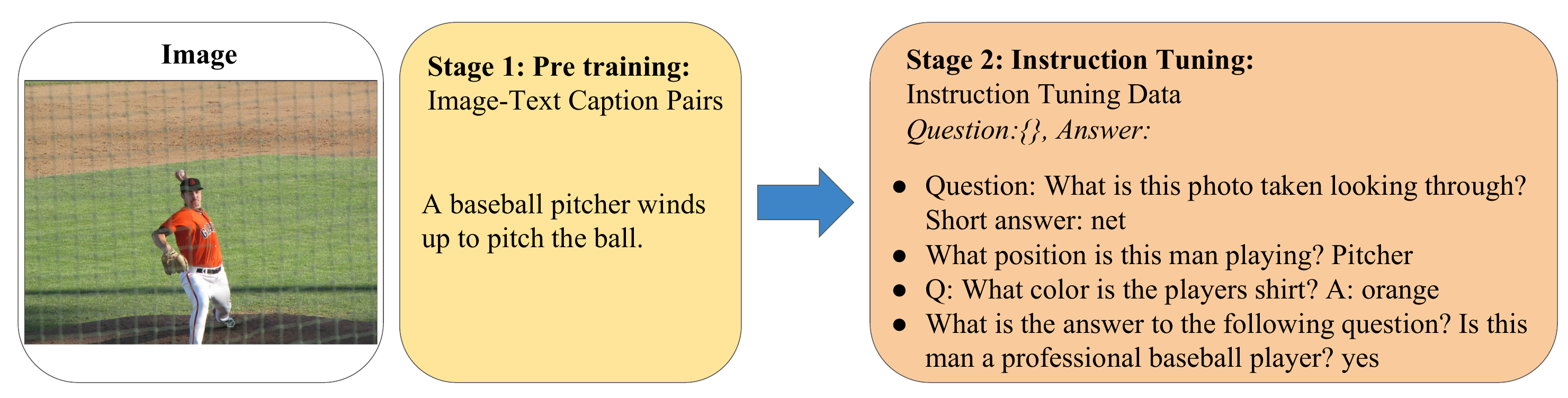}
\caption{\textbf{A typical multi-stage VLM training paradigm.} The training process involves two key stages. For Q-former, the first stage is done by \citep{BLIP-2} where image and text caption pairs are pre-trained to accomplish a raw alignment between visual and language modalities. As for the patch feature, we followed \citep{LLaVA} to use the same pre-training dataset. In the second stage, the alignment is further refined using instruction tuning VQA data, which facilitates a more detailed understanding of visual input based on language instructions.}
\label{fig:twostage}
\end{figure}
We adopted the typical two-stage training scheme: 1) In the pre-training stage, the goal is to align the LLM with visual information using image-text pairs from image captioning datasets that provide global descriptions of images. 2) After pre-training, the LLM becomes familiar with the visual embedding space and can generate descriptions of images. However, it still lacks the capability to discern the finer details of images and respond to human questions. In the second stage, we use instruction tuning data to enhance performance and further align the visual embeddings with the LLM and human values. Recent methods have predominantly adopted a two-stage training approach~\citep{MiniGPt4, LLaVA, mPlug-OwL} except PandaGPT~\citep{pandagpt}, which utilizes a one-stage training method, has also demonstrated commendable results. 
% It directly performs instruction tuning on the projection layer and LLM together, i.e., no pre-training alignment of the patch feature projection. 
In BLIVA, our visual assistant branch, specifically the encoded patch embeddings, diverges from the approach of BLIP-2~\citep{BLIP-2}, which uses a 129M pre-training dataset. Instead, it leverages a more compact 0.5M pre-training caption data following~\cite{LLaVA}. This presents a more efficient strategy for aligning the visual encoder and LLM at the first stage. We employed language model loss as our training objective. The model learns to generate subsequent tokens based on the preceding context.
% In our experiment, we test both one-stage and two-stage training and the results demonstrate that two-stage training leads to better performance.

% The primary objective of the training process is to maximize the log-likelihood of the tokens. It is important to note that we mask the visual embeddings and use only discrete text tokens in the calculation of the training loss.

\subsection{Thumbnails Dataset}
\begin{figure*}[htbp]
\begin{center}
\includegraphics[scale=0.7, width=0.9\textwidth]{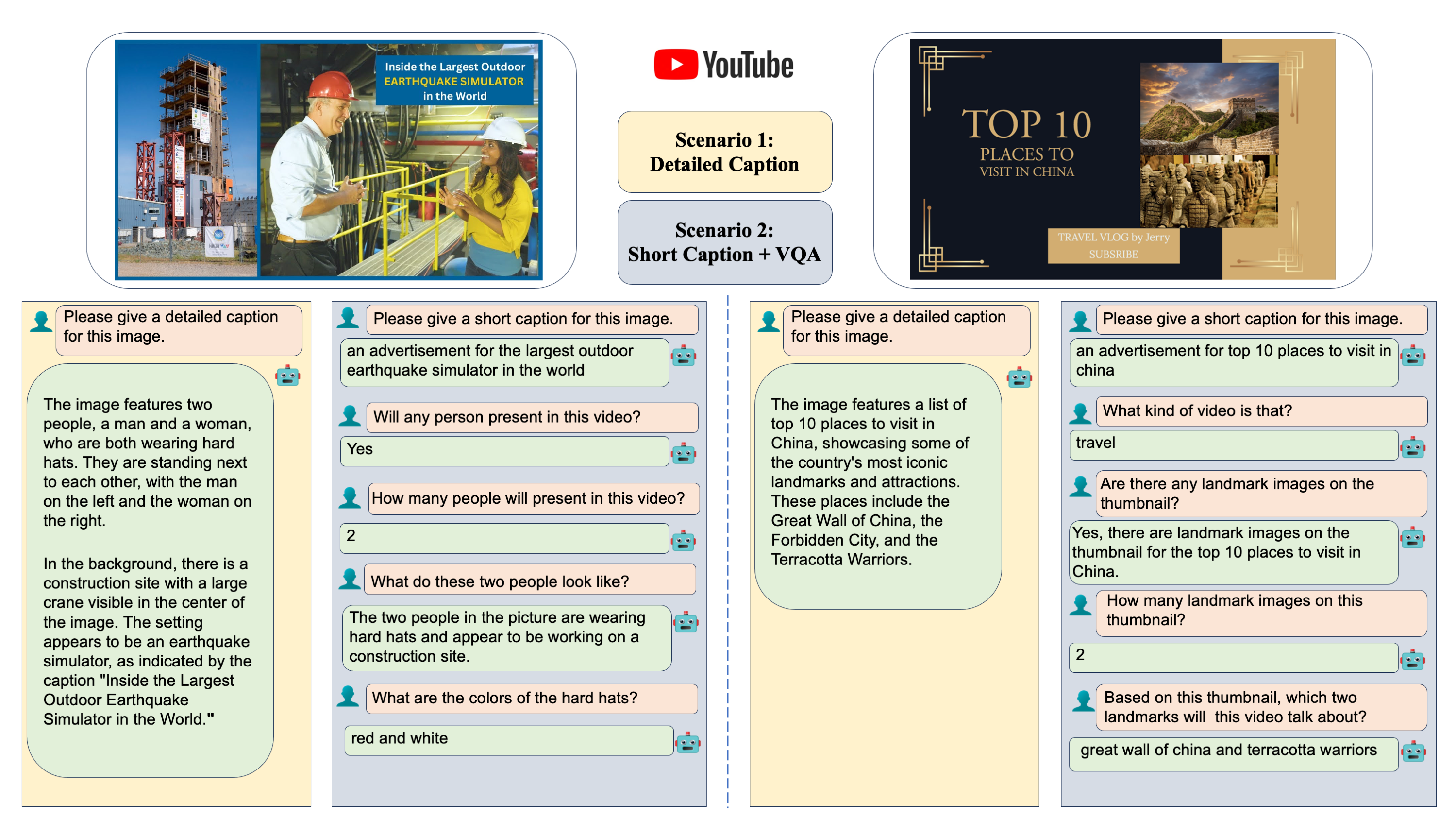}
\end{center}
% \vspace{-0.1in}
\caption{\textbf{Two Sample Scenarios from the YTTB-VQA Dataset}. This dataset demonstrates the dual application of BLIVA. The first scenario highlights BLIVA's capability to provide detailed captions that encompass all visual information within an image. The second scenario showcases BLIVA's utility in summarizing visual data into concise captions, followed by its ability to field more detailed visual queries posed by users.}
\label{fig:youtube_example}
\end{figure*}

To showcase the wide-ranging industry applications made feasible by BLIVA, we assess the model by introducing a new evaluation dataset, named \textbf{YTTB-VQA} which consists of 400 \textbf{Y}ou\textbf{T}ube \textbf{T}hum\textbf{b}nail \textbf{V}isual \textbf{Q}uestion-\textbf{A}nswer pairs to evaluate the visual perception abilities of in-text images. It covers 11 different categories which is illustrated in the Appendix Figure \ref{sec:YTTB chart}. During the data collection, we randomly selected YouTube videos with text-rich thumbnails from different categories. We recorded the unique video ID for each YouTube video and obtained the high-resolution thumbnail from the URL "http://img.youtube.com/vi/\textless YouTube-Video-ID\textgreater/maxresdefault.jpg". After retrieving all the YouTube thumbnails, we created the annotation file with the following fields: "video\_id" representing the unique identification for a specific YouTube video, "question" representing the human-made question based on the text and image in the thumbnail, "video\_classes" representing the 11 video categories, "answers" representing the ground truth answer, and "video\_link" representing the URL link for each YouTube video. Our Youtube thumbnail datasets are available at \textcolor{blue}{\url{https://huggingface.co/datasets/mlpc-lab/YTTB-VQA}}. 

We also provide two sample scenarios from the YTTB-VQA dataset. Figure~\ref{fig:youtube_example} illustrates BLIVA's capability to provide detailed captions and answer users' visual questions.

\section{Experiment}

\label{sec:exp}
In this section, we conduct extensive experiments and analyses to show the efficacy of our model. We evaluate our model, baseline, and other SOTA models on 10 OCR-related tasks and 8 general {(not particularly text-rich)} VQA benchmarks, including image captioning, image question answering, visual reasoning, visual conversational QA, image classification, and video question answering. We also evaluated on a comprehensive multimodal LLM benchmark (MME). We seek to answer the following:
%knowledge-grounded image question answering

\begin{itemize}
    \setlength\itemsep{-0.em}
    \item How does our proposed method compare to alternative single image embeddings approaches in text-rich VQA,  general VQA benchmarks and MME benchmark?  
    \item How do the individual components of our method influence its success?
    \item How does BLIVA enhance the recognition of YouTube thumbnails? 
\end{itemize}  

\subsection{Datasets}

 To demonstrate the effectiveness of patch embeddings, we followed \citep{insblip} to use the same training and evaluation data unless mentioned explicitly. The detailed dataset information can be found at Appendix~\ref{appendix:dataset}. 
 % we also follow the exact same training setting including hyperparameters, training datasets balancing scheme, and inference methods~\citep{insblip}.

 % Following \citep{insblip}, we also balance different sizes of each training dataset by square root smoothing. Given $\displaystyle D$ dataset with sizes $\displaystyle \{ S_1, S_2, \dots,  S_D \}$, the probability of a data sample being selected from a dataset $\displaystyle d$ is formulated as $\displaystyle p_d  = \frac{\sqrt{S_d}}{\sum_{i=1}^D \sqrt{S_i}}$. This prevented the model from overfitting on smaller datasets while underfitting on larger datasets.

\begin{table*}[ht!]
\begin{center}
%0.85
\resizebox{0.98\textwidth}{!}{%

\begin{tabular}{l| ccccccccccccc c}
\hline
%\multicolumn{2}{c}
   & STVQA $\uparrow$  & OCRVQA $\uparrow$  & TextVQA $\uparrow$ & DocVQA $\uparrow$& InfoVQA $\uparrow$ & ChartQA $\uparrow$  & {ESTVQA} $\uparrow$& FUNSD $\uparrow$   & SROIE $\uparrow$ & POIE $\uparrow$&  \colorgray Average $\uparrow$ \\ 
   \hline 
OpenFlamingo~\citep{openflamingoanas_awadalla_2023_7733589}           & 19.32 & 27.82 & 29.08  & 5.05 & 14.99 & 9.12 &  {28.20}        & 0.85   & 0.12  & 2.12                &  \colorgray 13.67  \\
BLIP2-OPT\textsubscript{6.7b}~\citep{BLIP-2}                   & 13.36 & 10.58 & 21.18   &  0.82 & 8.82 & 7.44 &  {27.02}     & 0.00   & 0.00  & 0.02   &  \colorgray 8.92     \\
BLIP2-FLanT5\textsubscript{XXL}~\citep{BLIP-2}                & 21.38 & 30.28 & 30.62   &  4.00 & 10.17 &  7.20  &  42.46        & 1.19 & 0.20  & 2.52 &  \colorgray 15.00\\
MiniGPT4~\citep{MiniGPt4}                 & 14.02 & 11.52 & 18.72   & 2.97  & 13.32 &  4.32  &  {28.36}        & 1.19  & 0.04  & 1.31        &   \colorgray9.58                     \\
LLaVA~\citep{LLaVA}                    & 22.93 & 15.02 & 28.30   &  4.40& 13.78 &   7.28  &  {33.48}       & 1.02   & 0.12  & 2.09                    &   \colorgray 12.84                    \\

mPLUG-Owl~\citep{mPlug-OwL}               & {26.32} & 35.00 & 37.44   &  6.17& \textbf{16.46} &  \textbf{9.52}  &   \textbf{\textbf{49.68}}       & 1.02   & 0.64  & \textbf{3.26}                &  \colorgray 18.56                     \\ 
InstructBLIP (FlanT5\textsubscript{XXL})~\citep{insblip} &26.22 & 55.04 &36.86 &4.94 &10.14  &8.16 & 43.84 &1.36 &0.50 & 1.91 &  \colorgray 18.90 \\
InstructBLIP (Vicuna-7B)~\citep{insblip}     &28.64 &47.62 &39.60 &5.89 & 13.10 & 5.52 & 47.66& 0.85& 0.64&  2.66  &  \colorgray 19.22 \\

\hline
 \modelName (FlanT5\textsubscript{XXL})  &{28.24} & 61.34 &39.36 &5.22 & 10.82 & 9.28 & 45.66 & \textbf{1.53} & 0.50 & 2.39 &  \colorgray 20.43 \\
 BLIVA (Vicuna-7B) & \textbf{29.08} & \textbf{65.38}  & \textbf{42.18} & \textbf{6.24}  &13.50 & 8.16& {48.14}  &1.02 & \textbf{0.88}& 2.91 &  \colorgray \textbf{21.75}\\ 
% BLIVA (Vicuna-7B) w/o pre-training & \textbf{28.86}  &\textbf{65.04}  & \textbf{40.7}  & \textbf{6.65} &14.28 & 8.24 & 47.72 &  1.19 & \textbf{1.66} &2.83 & \textbf{21.72}\\

\end{tabular}%
}
\end{center}
% \vspace{-0.07in}
\caption{\textbf{Zero-Shot OCR-Free Results on Text-Rich VQA benchmarks}. This table presents the accuracy (\%) results for OCR-free methods, implying no OCR-tokens were used. Note that our work follows InstructBLIP which incorporated OCR-VQA in its training dataset, thus inevitably making OCR-VQA evaluation not zero-shot.}
% Except for InstructBLIP, all other results are sourced from \citep{liu2023hidden} on May 20th, 2023, with which we ensure consistency in our results.}
\label{tab:table2}
\end{table*}
 % SoTA metrics employ Average Normalized Levenshtein Similarity (ANLS) for ST-VQA and DocVQA, while accuracy is used for TextVQA and OCR-VQA,

\subsection{Implementation Details} 

We selected the ViT-G/14 from EVA-CLIP~\citep{sun2023evaclip} as our visual encoder. In line with InstructBLIP, we employed Vicuna-7B which has been instruction-tuned from LLaMA~\citep{touvron2023llama} and serves as our LLM. Additional details can be found in Appendix~\ref{implemnt details}.

% \textbf{Baselines} In this work, we select Instructed BLIP as our baseline, which is the most parameter-efficient and powerful model, and demonstrate the effectiveness of our proposed pipeline

\subsection{Results \& Discussions}

\label{results}

We introduce our results in the context of each of our three questions and discuss our main findings.

\emph{\textbf{1.~~~How does our proposed method compare to alternative single image embeddings approaches in text-rich VQA,  general VQA benchmarks and MME benchmark? }}

\begin{table*}[ht!]
\begin{center}
%\begin{tabular}{l|c c c c c c c |c c c}
%0.75
 \resizebox{0.95\textwidth}{!}{%
\begin{tabular}{l|c c c c c |c c c}
 \hline
Models         &VSR $\uparrow$&IconQA $\uparrow$&TextVQA $\uparrow$ &Visdial $\uparrow$ &Flickr30K  $\uparrow$&HM $\uparrow$ &VizWiz  $\uparrow$ &MSRVTT  $\uparrow$  \\
& & &  & & &(val)  &(val-dev)  &(val-dev) \\
\hline
Flamingo-3B~\citep{Flamingo}           & - & - & 30.1 & -  & 60.6& -& - & -  \\
Flamingo-9B~\citep{Flamingo}               & - & - & 31.8 & -  & 61.5& - & - & -   \\
Flamingo-80B~\citep{Flamingo}                 & - & - & 35.0 & - & 67.2& - & - & -    \\
MiniGPT-4~\citep{MiniGPt4}       &50.65 & - & 18.56  & -& - & 29.0 & 34.78 & -   \\
%MiniGPT-4~\citep{MiniGPt4}       &50.65 &  7.89 & 18.56  & - & 29.0 & 34.78 & 5.65 & - & 8.4 & 6.3 \\
LLaVA~\citep{LLaVA} & 56.3 & - & 37.98 &  -  & -& 9.2 & 36.74  & -   \\
BLIP-2 (Vicuna-7B)~\citep{insblip}           &50.0 &39.7 & 40.1 & 44.9 &74.9 &50.2 & \textbf{49.34}  & 4.17 \\ 
InstructBLIP (Vicuna-7B)~\citep{insblip}     &54.3 &43.1 &50.1 &45.2 &82.4 &54.8 &43.3  &18.7  \\
\hline

InstructBLIP Baseline (Vicuna-7B)  &58.67   & 44.34 & 37.58 & 40.58&84.61& 50.6 & 44.10  & 20.97 \\

  \modelName (Vicuna-7B)                    &  \textbf{62.2} &  \textbf{44.88}&  \textbf{57.96} & \textbf{45.63} &  \textbf{87.1} & \textbf{55.6} &  42.9  &  \textbf{23.81} \\ % 54.94 HM 
\end{tabular}%
}
\end{center}
% \vspace{-0.07in}
\caption{\textbf{Zero-shot results on general {\em(not particularly text-rich)} VQA benchmarks}. Our baseline is obtained by directly finetuning InstructBLIP \citep{insblip}. For the three datasets on the right, due to the unavailability of test-set answers, we have evaluated them using validation dev.
%We were unable to download the LAION dataset due to it contains sensitive content, resulting in a lack of this training dataset.
Here, Visdial and HM denote the Visual Dialog and Hateful Memes datasets, respectively. Following previous works~\citep{Flamingo,ivqa,visdial_pretrain}, we report the CIDEr score~\citep{cider} for Flickr30K, AUC score for Hateful Memes, and Mean Reciprocal Rank (MRR) for Visual Dialog. For all remaining datasets, we report the top-1 accuracy (\%). Notably, for Text-VQA, we have followed InstructBLIP's method of using OCR-tokens for comparison. While InstructBLIP also included GQA, iVQA, and MSVDQA, we were unable to access these datasets due to either unresponsive authors or the datasets being removed from their websites. For ScienceQA and Nocaps, we were unable to reproduce the results of InstructBLIP, hence their results are not reported here.}
\label{tab:table1}
\end{table*}

\textbf{Zero-shot evaluation for text-rich VQA benchmarks}
We compared our data with state-of-the-art Multimodality LLMs. This includes LLaVA, which showcases robust OCR capabilities using only patch embedding. 
% and its recent OCR-centric version, LLaVAR, which utilizes additional OCR training data.
We also considered BLIP2's previous best version, BLIP-FLanT5xxL, the state-of-the-art vision-language model mPlug-Owl (trained on a vast amount of both text and vision-text data), and our baseline, InstructBLIP. The results are illustrated in Table~\ref{tab:table2}. Our model consistently shows significant improvement across all the text-rich VQA datasets compared to InstructBLIP. Note that since InstructBLIP utilized OCR-VQA as its training dataset, the comparison for this specific dataset isn't zero-shot. We evaluated both InstructBLIP and our model using the OCR-VQA validation set. BLIVA achieved state-of-the-art results among 6 text-rich datasets while mPlug-Owl performed the best in 4 datasets. Compared to mPlug-Owl, which employs about 1104M image captioning data in the Pre-training stage, BLIVA only employs 558K image caption pairs which could explain why BLIVA is not performing the best in information-based VQA tasks such as InfoVQA, ChartQA and ESTVQA. BLIVA demonstrates the best performance on average compared to all previous methods, underscoring our design choice to employ learned query embeddings, further aided by encoded patch embeddings.
% While LLaVAR benefited from a newly collected OCR dataset consisting of 422K text-rich images, OCR-VQA was the only OCR-related data we used during training. Our performance surpassed LLaVAR except for ST-VQA, underscoring our design choice to employ learned query embeddings, further aided by encoded patch embeddings. 
% Note here all the results are only based on the image and question with no ocr token used, which meets the real-life scenario the best. 

\begin{table*}[ht!]
\begin{center}
%0.75
 \resizebox{0.95\textwidth}{!}{%
%\begin{tabular}{l|c c c c c c c |c c c}
\begin{tabular}{l|c c c c c |c c c}
 \hline
Models         &VSR $\uparrow$&IconQA $\uparrow$&TextVQA $\uparrow$ &Visdial $\uparrow$ &Flickr30K  $\uparrow$&HM $\uparrow$ &VizWiz  $\uparrow$ &MSRVTT  $\uparrow$  \\
& & &  & & &(val)  &(val-dev)  &(val-dev) \\
\hline
BLIP-2 (FlanT5\textsubscript{XXL})~\citep{BLIP-2}          &68.2 &45.4 & 44.1 & 46.9&73.7 &52.0 & 29.4  & 17.4  \\ 
InstructBLIP (FlanT5\textsubscript{XXL})~\citep{insblip} &65.6 &51.2 &46.6 &\textbf{48.5}  &83.5 &\textbf{53.6} &41.35 &20.79 \\
\hline
\modelName (FlanT5\textsubscript{XXL})   & \textbf{68.82} & \textbf{52.42} &\textbf{57.2}& 36.18 & \textbf{87.66}&  50.0 & \textbf{43.97} &   \textbf{23.78} 
\end{tabular}%
}
\end{center}
% \vspace{-0.05in}
\caption{\textbf{Zero-shot results on general {\em(not particularly text-rich)} VQA benchmarks for models with open LLM eligible for commercial use}. Here, the commercial use applicable LLM we reported is FlanT5\textsubscript{XXL}. Same as Table~\ref{tab:table1}, we report the same evaluation datasets with the same evaluation metrics.   }
\label{tab:flan}
\end{table*}

\begin{table*}[ht!]
\begin{center}
%0.8
\resizebox{0.95\textwidth}{!}{%
\begin{tabular}{l|c|cccccccccc|cccc|c}
\hline

Model & Overall $\uparrow$ &\multicolumn{10}{c}{Perception $\uparrow$ } & \multicolumn{4}{c}{Cognition $\uparrow$ } & Avg. $\uparrow$  \\
  &  & Exist. & Count & Pos. & Color & OCR & Poster & Cele. & Scene & Land. & Art. & Comm. & NumCal. & Trans. & Code &  \\
\hline
LLaVA\cite{LLaVA} & 712.5 & 50.0 & 50.0 & 50.0 & 50.0 & 50.0 & 50.0 & 48.8 & 50.0 & 50.0 & 49.0 & 57.1 & 50.0 & 57.5 & 50.0 & 50.9  \\

MiniGPT-4\cite{MiniGPt4} & 694.3 & 68.3 & 55.0 & 43.3 & 43.3 & 57.5 & 41.8 & 54.4 & 71.8 & 54.0 & 60.5 & 59.3 & 45.0 & 0.0 & 40.0  & 49.6 \\

mPLUG-Owl\cite{mPlug-OwL} & 1238.4 & 120.0 & 50.0 & 50.0 & 50.0 & 65.0 & 136.1 & 100.3 & 135.5 & \underline{159.3} & 96.3 & 78.6 & \underline{60.0} & \underline{80.0} & 57.5 & 88.5 \\

InstructBLIP\cite{insblip} & 1417.9 & \underline{185.0} & \underline{143.3} & 66.7 & 66.7 & 72.5 & 123.8 & 101.2 & \underline{153.0} & 79.8 & \underline{134.3} & \underline{129.3} & 40.0 & 65.0 & 57.5 & 101.3\\

BLIP-2\cite{BLIP-2} & 1508.8 & 160.0 & 135.0 & \underline{73.3} & \underline{73.3} & \underline{110.0} & \underline{141.8} & \underline{105.6} & 145.3 & \underline{138.0} & \underline{136.5} & 110.0 & 40.0 & 65.0 & \underline{75.0} & 107.8\\

\hline
\rowcolor{gray!20} BLIVA  & \textbf{1669.2} & \underline{180.0} & \underline{138.3} & \underline{81.7} & \underline{180.0} & \underline{87.5}& \underline{155.1} & \underline{140.9} & \underline{151.5} & 89.5 & 133.3  & \underline{136.4} & \underline{57.5} & \underline{77.5} & \underline{60.0} & \textbf{119.2} \\

\end{tabular}%\
}
\end{center}
\caption{
\textbf{Evaluation of MME-Benchmark}. Here we report the results on all the sub tasks, including Existence(Exist.), Count, Position(Pos.), Color, OCR, Poster, Celebrity(Cele.), Scene, Landmark(Land.), Artwork(Art.), Commonsense Reasoning(Comm.), Numerical Calculation(NumCal.), Text Translation(Trans.), Code Reasoning(Code) and the task-level average (Avg.). We \textbf{bold} the highest overall and average score and highlight the Top-2 model of each sub task with \underline{underline}. }
% \vspace{1.5 pt}
% \vspace{-0.05in}
\label{table:mme}
\end{table*}

\begin{table*}[ht!]
\begin{center}
%0.8
\resizebox{0.95\textwidth}{!}{%
\begin{tabular}{l l l l l| c c c c c c c c c c | g} %{l l l l| l l l l l l l l l l }
 \hline
 InstructBLIP   & Baseline  & Patch &  Pre- & Finetuning & ST- &OCR- &Text-- &Doc-  &  Info- & Chart- & EST- & FUNSD & SROIE & POIE & Improvement\\
 \citep{insblip} & (Instruction & Embedding & Training & LLM & VQA &VQA &VQA &VQA  &  VQA & QA & VQA &  &  &  &\\
  & Tuning  & & & & & & & &  & & & & & &  \\
  & Qformer)   & & & & & & & &  & & & & & & \\
%  & \textcolor{blue}{Finetuning }  &  \textcolor{red}{Fintuning Patch} &  \textcolor{red}{Fintuning Patch} & &&&&&&&& &&Average \\
 
%   &  & &  &ST-VQA &OCR-VQA &TextVQA &DocVQA  &  InfoVQA & ChartQA & ESTVQA & FUNSD & SROIE & POIE & Improvement  \\
  
% & & \textcolor{red}{Params Initialized} &\textcolor{red}{Params Initialized}  & &&&&&&&& & &\\
 
\hline
\checkmark &  & & &  &28.64 &47.62 &39.60 &5.89 & 13.10 & 5.52 & 47.66& 0.85& 0.64&  2.66 & + 0 \% \\
\checkmark &\checkmark& &    & &\textbf{30.08} &65.8  &40.5  & 6.13 & 12.03 &8.08 &47.02  &0.85 &0.57 & 2.62 & + 7.40\%\\
\checkmark &  \checkmark & \checkmark & & & 28.86  &65.04  & 40.7  & \textbf{6.65} &\textbf{14.28} & \textbf{8.24} & 47.72 & \textbf{1.19} & \textbf{1.66} &2.83 & + \textbf{31.72\%} \\
\checkmark &  \checkmark & \checkmark  & \checkmark  & & 29.08 & 65.38  & \textbf{42.18} & 6.24  &13.50 & 8.16& \textbf{48.14} &1.02 & 0.88& \textbf{2.91} & + 17.01\% \\ 
% \checkmark & \checkmark & &\checkmark & \checkmark & 29.94  & \textbf{66.48} & 41.9  & 6.47  \\
% this is finetuing LLM results
% \rowcolor{mygray}
% \multicolumn{4}{l}{BLIVA with Pre-training + Finetuning LLM} 
\checkmark & \checkmark &\checkmark &\checkmark & \checkmark & 29.94  & \textbf{66.48} & 41.9  & 6.47  & 12.51 & 7.52  & 46.76 & 1.02 & 0.51 & 2.85 & + 9.65\%
\\
\end{tabular}%\
}
\end{center}
% \vspace{-0.05in}
\caption{\textbf{Results of adding individual techniques of our framework in text-rich VQA benchmarks}. We include four ablations that accumulate each technique (i) baseline: instruction tuning InstructBLIP's Qformer. (ii) instruction tuning patch embeddings  (iii) pre-training stage of patch embeddings (iv) Finetuning LLM with LORA during the instruction tuning stage.}
\label{tab:ablation_text}
\end{table*}

\begin{table*}[ht!]
\begin{center}
%\begin{tabular}{l|c c c c c c c | c c c}
%0.8
\resizebox{0.95\textwidth}{!}{%
\begin{tabular}{l l l l l|ccccc|ccc|g }
 \hline
InstructBLIP  &   Baseline & Patch  & Pre- & Finetuning         &VSR &IconQA &TextVQA &Visdial  &Flickr  &HM  &VizWiz  &MSRVTT & Improvement  \\
 ~\citep{insblip} & (Instruction   &  Embeddings & Training &  LLM        & & & &  & 30K & (val) & (val-dev) & (val-dev) & \\
  & Tuning  & & & & & & & &  & & && \\
  & Qformer)  & & &  & & & & &  & && &  \\
\hline
%\modelName (Vicuna-7B)\textsubscript{$\displaystyle 384^2$}       \\
% \modelName (Vicuna-7B)                      &62.2 &44.88 &57.96 & 45.63 &54.94 &42.9 &58.11 &23.81 &108.2 &87.1 \\
% - Pretraining stage  &58.85 &44.91 & 58.8 & 41.67 & 49.05 &42.83 & 54.49 & 23.70 & 107.4 & 87.4\\
% - patch embeddings  & 57.7 &44.80 &52.62 &40.44 & 52.8& 43.7 & 56.82 &22.09 & 107.7 & 87.4 \\
% + LoRA LLM             & 51.39 & 41.34 &57.82 & 42.32  &46.19 & 44.91 & 50.02 & 22.67 & 99.81 & 82.7 \\        

%&& 57.7 &44.80 &52.62 &40.44 & 87.4 & 52.8& 43.7  &22.09  \\
\checkmark &  & \ &\ &  \    &54.3 &43.1 &50.1 &45.2 &82.4 &54.8 &43.3  &18.7  & + 0\% \\
\checkmark & \checkmark & \ &\ &  \ &  58.67   & 44.34 & 37.58 & 40.58&84.61& 50.6 & 44.1  & 20.97 & - 1.91\% \\ 
%&54.3 &43.1 &50.1 &45.2 &6.26 &43.3&60.5 &18.7 &123.1 &82.4\\
%& 57.7 &44.80 &52.62 &40.44 & 52.8& 43.7 & 56.82 &22.09 & 107.7 & 87.4 \\
\checkmark &    \checkmark & \checkmark &\ &   &58.85 &\textbf{44.91} & \textbf{58.8} & 41.67  & \textbf{87.4}& 49.1 &42.83 & 23.70 & + 5.43\% \\
\checkmark & \checkmark &  \checkmark & \checkmark &    & \textbf{62.2} & 44.88 & 57.96 &  \textbf{45.63} & 87.1 & \textbf{55.6} & 42.9 & \textbf{23.81} & \textbf{+ 8.61\%} \\
  \checkmark &  \checkmark & \checkmark &\checkmark & \checkmark   & 51.39 & 41.34 &57.82 & 42.32  & 82.7 &46.2 & \textbf{44.91}  & 22.67 & + 1.15\% \\

\end{tabular}%\
 }
\end{center}
\caption{\textbf{Results of adding individual techniques of our framework in general {\em(not particularly text-rich)}  VQA benchmarks}. We include  four ablations that accumulate each technique same as in Table~\ref{tab:ablation_text}.}
\label{tab:ablation_general}
\end{table*}

\textbf{Zero-shot evaluation for general {\em(not particularly text-rich)} VQA benchmarks} Next, we compared BLIVA with models that employ single image features. Results are given in Table~\ref{tab:table1} and in Table~\ref{tab:flan} for LLMs available for commercial use. Our model consistently and significantly outperformed the original InstructBLIP model in VSR, IconQA, TextVQA, Visual Dialog, Hateful Memes, MSRVTT, and Flickr30K. For VizWiz, our model nearly matched InstructBLIP's performance. This naturally raises the question: why didn't additional visual assistance improve all the benchmarks? We speculate that the additional visual information didn't aid VizWiz task. We continue to investigate this phenomenon in the next ablation study section. Overall, our design not only achieved significant improvements in understanding text-rich images but also improves 7 out of 8 general VQA benchmarks.

\textbf{MME Benchmark} We further evaluated BLIVA on a comprehensive Multlimodal LLM benchmark (MME)~\citep{fu2023mme}. As illustrated in Table \ref{table:mme}, BLIVA demonstrates the best performance among the current methods on average for both the perception and cognition tasks. For all text-rich tasks such as OCR, Poster, Numerical Calculation, Text Translation, and code, BLIVA outperforms InstructBLIP. BLIVA achieved top 2 performance across all the tasks except artwork and landmark which demand extensive informational knowledge.  This is consistent with our findings from informational VQA, indicating that our light-weight pre-training stage and the missing LAION-115M web image caption dataset during instruction tuning stage both likely contribute to a degradation in BLIVA's internet knowledge base. 

%This result could be caused by the fact that we lack LAION-115M image caption training data due to the illegal information contaminated this dataset.
%todo add more models for this table

\emph{\textbf{{2.~~~How do the individual components of our method influence its success?}   }} %\vspace{-0.01in}  

To investigate the impact of image-encoded patch embeddings, the pre-training stage, and fine-tuning the LLM, we conducted ablation studies, incorporating each element respectively. For simplicity, here we only conduct ablation on the \modelName (Vicuna-7B) model. Since our baseline is the InstructBLIP, we report the results of using baseline alone as directly finetuning the InstructBLIP model with our data and implementation.

% \textbf{Ablation in text-rich VQA benchmarks} \ For text-rich image related tasks, Table~\ref{tab:ablation_text} illustrates the results of adding each technique separately. Without the pre-training stage, ST-VQA, OCR-VQA, and TextVQA all showed degrades in their performance except DocVQA, indicating the importance of image-text pair data for global general alignment. Since all models perform very poorly in this task since DocVQA consists of documents with very tiny text characters when the image resolution is down-scaled to 224. This discrepancy in performance is then not significant. When the encoded patch embeddings are removed, we observe all the tasks' performance downgrades, which indicating our design of employing patch embeddings provides more detailed visual information. It also matches our hypothesis that an additional visual assistant is responsible for improving the visual knowledge where the query embeddings neglect or are limited to extract. 

\textbf{Ablation in text-rich VQA benchmarks} \ For text-rich image related tasks, Table~\ref{tab:ablation_text} illustrates the results of adding each technique separately. Compared to the baseline, adding patch embeddings improved performance across all tasks with the exception of ST-VQA and OCR-VQA. This can stem from data contamination, as STVQA includes data already present in InstructBLIP’s Qformer training set but not included in patch embedding’s training set. 
% We conjecture that ST-VQA emphasizes high-level semantic information in images, making it better suited for fine-tuning query embeddings in isolation. 
Without the pre-training stage, the performance of ST-VQA, OCR-VQA, TextVQA, ESTVQA, and POIE decreased, while the rest are benefited. Since the pre-training stage employs image caption pairs, we observed that it didn't benefit BLIVA’s performance in text-rich VQA tasks as consistently as in the general VQA tasks. Considering the improvement of all tasks, pre-training is still adopted. 
% The discrepancy can be attributed to the fact that all models struggle with DocVQA. This is largely due to the dataset containing documents with minuscule text characters, which become further illegible when the image resolution is downscaled to 224.
BLIVA on average
outperforms InstructBLIP by 31.72\% without pre-training and 17.01\% with it, both outpacing the 7.40\% improvement from instruction tuning Qformer.
These studies indicate that our design of employing patch embeddings provides more detailed visual information. It also supports our hypothesis that an additional visual assistant improves visual knowledge in areas where the query embeddings either neglect or have limited extraction capabilities.

\textbf{Ablation in general {\em(not particularly text-rich)} VQA benchmarks} \  As illustrated in Table~\ref{tab:ablation_general}, 
the presence of encoded patch embeddings improves performance in all benchmarks significantly except HM and VizWiz. For tasks where we observed a drop in performance, such as HM, which focuses on interpreting the feeling of hatefulness, and VizWiz, which predicts whether a visual question can be answered. We conjecture these tasks can be fulfilled by utilizing global-level query embeddings information such as feeling the hatefulness in the image or if the image's object is unrelated to the question asking. When adding the first pre-training stage, the performance for VSR, VisDial, HM, and  MSRVTT tasks improves substantially while others are kept roughly the same. These ablation results confirmed the necessity of two-stage training. 
%the visual embeddings for an initial alignment with LLM and then instruction tuning the connection module to enhance the alignment for better performance. 
During the instruction tuning stage, we also experimented with fine-tuning the LLM using LoRA in conjunction with Q-former and encoded patch embeddings. However, this approach didn't yield as much improvement as our best model and even reduced performance in many tasks. Nonetheless, we have included these results in the ablation study for completeness. We conjecture that frozen LLM has a satisfactory understanding of visual information after our two-stage alignment. The visual embeddings are interpreted as a "foreign language" to LLM and thus finetuning LLM together is not needed in this case. 

\emph{\textbf{{3.~~~How does BLIVA enhance the recognition of YouTube thumbnails?} }} %\vspace{-0.01in}

\textbf{Youtube Thumbnails Evaluation} \
Table~\ref{tab:youtube} illustrates the results of the youtube thumbnail dataset with BLIVA achieving the best performance. From an application perspective, BLIVA has the ability to extract extra visual information from images besides extracting information from YouTube captions alone like LLMs. Our success in this use case can be further expanded to large-scale thumbnail images.

\begin{table}[ht!]
\begin{center}
\resizebox{0.9\columnwidth}{!}{%
\begin{tabular}{l|l }
 \hline
Models        & Accuracy (\%) \\
\hline
MiniGPT4~\citep{MiniGPt4} & 47.75
\\ LLaVA~\citep{LLaVA} & 41.75  \\ 
InstructBLIP (Vicuna-7B)~\citep{insblip}  & 82.2 \\
\hline
 \modelName (Vicuna-7B) & \textbf{83.5} \\
%\modelName (Vicuna-7B) Text Expert \\
% \modelName (FlanT5  88.0 \\

% BLIP2-OPT\textsubscript{6.7b}
% BLIP2-FLanT5
% LLaVAR (224)~\citep{LLaVAR}   
% mPLUG-Owl~\citep{mPlug-OwL}   
% InstructBLIP (FlanT5\te

%  \modelName (Vicuna-7B)
%59.04(ocr)
\end{tabular}
}
\end{center}
\caption{\textbf{Evaluation results of our collected Youtube thumbnails dataset}. We report the top-1 accuracy (\%).}
\label{tab:youtube}
\end{table}
% \vspace{-0.15in}

\subsection{Qualitative Analysis}
We use real-life scene images, movie posters, webpages, and memes to demonstrate our model's performance regarding interaction with humans based on text-rich images. The examples are in Appendix~\ref{more examples}. BLIVA showcases exceptional OCR capabilities, paired with a robust localization ability that accurately identifies texts and objects within images.  

%BLIVA’s reply is strictly based on visual content without hallucination like InstructBLIP. 

\section{Conclusion}

In this paper, we illustrate the effectiveness of assisting learned query embeddings with encoded image patch embeddings as a visual assistant. This straightforward yet innovative design bolsters performance in both general VQA benchmarks and text-rich VQA benchmarks. Our model, \modelName, demonstrates superior performance in both academic benchmarks and qualitative real-world examples. Moreover, human evaluation of the model's performance reveals that \modelName struggles with deciphering numerical symbols in images. This could be attributed to the reduced pixel representation often used for these symbols and needs future work to develop valuable insights.  
% Moreover, human evaluation of the model's performance reveals that \modelName still struggles with document-related commonsense reasoning, and creative generation which might be due to the reason of insufficient document-related datasets and need future work to develop valuable insights.  
Our work also demonstrates the effectiveness of mixing different types of visual embeddings. We encourage more future work to explore how to scale more visual embeddings to LLM which can be the key to the next stage of Large Vision-Language Models. 

\section{Acknowledgements}
\indent Zhuowen Tu is funded by NSF Award IIS-2127544.
% \textbf{Acknowledgements} \indent Zhuowen Tu is funded by NSF Award IIS-2127544.

\bibliography{aaai24}

\appendix
\newpage
\section{Appendix}
\subsection{Official Version}
Our official version in the AAAI proceedings will be notified and accessible at \textcolor{blue}{\url{https://github.com/mlpc-ucsd/BLIVA}}

\subsection{Time and Memory Complexity}
We report the training time as in one epoch and memory is on 8 NVIDIA A6000 with a batch size of 24. The inference time is based on NVIDIA A100. BLIVA outperforms InstructBLIP without sacrificing much computation power.

\begin{table}[ht!]
\begin{center}
\resizebox{0.95\columnwidth}{!}{%
\begin{tabular}{l l l l l}
 \hline
Models   & Training Time&Inference Time&Training Memory& Inference Memory \\
\hline 
InstructBLIP & 11.5h & 0.89s  & 34G & 18G\\
BLIVA (Ours)  & 14.8h & 0.91s & 40G  & 22G
\end{tabular}%
}
\end{center}
\caption{Comparison of Time and Memory Overheads}
\label{tab:speed}
\end{table}
\subsection{Text Fonts in images}
We evaluated BLIVA with two widely-used fonts, Times New Roman and Impact, in four colors—white, red, pale green, and blue—as detailed in Table \ref{tab:font}. We observed degradation of performance in Blue color. Since the background of image is gray-scale, the color contrast between image color and text color affects model's performance. BLIVA performs significantly better with light-colored fonts, like white or pale green, compared to darker colors like blue.

\begin{table}[ht!]
\begin{center}
\resizebox{0.95\columnwidth}{!}{%
\begin{tabular}{l l l l l}
\hline
Font Type / Color  & White & Red & Pale Green & Blue\\
\hline 
Times New Roman & 94.12\% & 94.12\% & 94.12\% & 58.82\% \\
Impact & 94.12\% & 88.24\% & 94.12\% & 70.59\% \\
%Caveat & 10/17 & 6/17 & 10/17 & 0/17 
\end{tabular}%
}
\end{center}
\caption{Comparison of performance across various fonts and colors, evaluated based on text-capture accuracy which is the number of words correctly detected over all the words.}
\label{tab:font}
\end{table}

\subsection{Encoding Methods in Visual Assistant branch}

We included different BLIVA variants focusing on encoding localized positional details in Table \ref{tab:differen encoding}. We picked four most representative datasets covering diverse tasks for this ablation study. BLIVA with linear projection outperforms other variants. A simple projector leads to better generalization.
\begin{table}[ht!]
\begin{center}
\resizebox{0.95\columnwidth}{!}{%
\begin{tabular}{l l l l l}
 \hline
Models        &Flickr30K &VSR &TextVQA &IconQA \\
\hline 
InstructBLIP &82.4 &54.3 &39.60 &43.1\\
 \textbf{BLIVA (Ours)}  &\textbf{87.1}  &\ 62.2   & \textbf{42.18} &  \textbf{44.88}  \\
\hline
+ Convolutional Position &85.87 &60.92 &41.28 &41.96 \\
+ ViT Block w/ Convolutional Position &86.59&\textbf{63.90}& 40.72  & 41.80\\
+ ViT Block w/ Relative Position & 84.57&60.31 & 41.24 & 41.96\\
+ MLP Layer & 82.9 & 59.57 & 42.12 & 43.79 \\

\end{tabular}%
}
\end{center}
\caption{BLIVA with different encoding methods }
\label{tab:differen encoding}
\end{table}

% Additional examples of our model's performance on various types of images.
\subsection{Datasets}
\label{appendix:dataset}
We followed \citep{insblip} to use the same training and evaluation data unless mentioned explicitly. Due to the illegal contents involved in LAION-115M dataset~\citep{schuhmann2021laion400m}, we cannot download it securely through the university internet. Besides lacking a subset of samples of image captioning, we keep all other training data the same. 
 It includes MSCOCO~\citep{lin2015microsoft} for image captioning, TextCaps~\citep{sidorov2020textcaps}, VQAv2~\citep{balanced_vqa_v2}, OKVQA~\citep{marino2019okvqa}, A-OKVQA~\citep{AOKVQA}, OCR-VQA~\citep{mishraICDAR19} and LLaVA-Instruct-150K~\citep{LLaVA}. 
 For evaluation datasets, we also follow \citep{insblip} but only keep Flickr30K~\citep{Flickr30K}, VSR~\citep{liu2023vsr}, IconQA~\citep{lu2022iconqa}, TextVQA~\citep{singh2019vqa}, Visual Dialog~\citep{visdial}, Hateful Memes~\citep{kiela2021hateful}, VizWiz~\citep{gurari2018vizwiz}, 
 %NoCaps~\citep{Agrawal_2019} ScienceQA~\citep{lu2022learn}
 and MSRVTT QA~\citep{msvdqa} datasets. Here, for Vizwiz, since there's no ground truth answer for the test split, we choose to use a validation split. For Hateful Memes, the test split also misses answers, so we picked the same number of examples from the training set as our evaluation data. InstructBLIP originally also had GQA~\citep{gqa} and iVQA~\citep{ivqa}; we contacted the authors for access to their datasets but received no reply yet. As for MSVDQA~\citep{msvdqa}, the authors completely removed this dataset from their competition website. For OCR task datasets, we followed~\citep{liu2023hidden} to select OCR-VQA~\citep{mishraICDAR19}, Text-VQA~\citep{singh2019vqa}, ST-VQA~\citep{stvqa}, DOC-VQA~\citep{docvqa}, InfoVQA~\citep{mathew2022infographicvqa}, ChartQA~\citep{masry2022chartqa}, ESTVQA~\citep{wang2020general}, FUNSD~\citep{jaume2019funsd}, SROIE~\citep{huang2019icdar2019}, and POIE~\citep{kuang2023visual}.

\subsection{Data Pre-processing}

We followed~\cite{insblip} to use the same data pre-processing methods, which are attached below. 

\begin{lstlisting}[language=Python]
class BlipImageTrainProcessor(BlipImageBaseProcessor):
    def __init__(
        self, image_size=224, mean=None, std=None, min_scale=0.5, max_scale=1.0
    ):
        super().__init__(mean=mean, std=std)

        self.transform = transforms.Compose(
            [transforms.RandomResizedCrop(
                image_size,
                scale=(min_scale, max_scale),
                interpolation=InterpolationMode.BICUBIC,
                ),
            transforms.RandomHorizontalFlip(),
            transforms.ToTensor(),
            self.normalize,
            ]
        )
    def __call__(self, item):
        return self.transform(item)
\end{lstlisting}

The first class \texttt{BlipImageTrainProcessor} is used to pre-process the images for the training purpose. Specifically, it randomly crops and resizes images to 224 * 224 with an interpolation method of Bicubic, possibly flips them horizontally, converts them to tensor format, and normalizes them using mean = (0.48145466, 0.4578275, 0.40821073) and standard deviation = (0.26862954, 0.26130258, 0.27577711).

\begin{lstlisting}[language=Python]
class BlipImageEvalProcessor(BlipImageBaseProcessor):
    def __init__(self, image_size=224, mean=None, std=None):
        super().__init__(mean=mean, std=std)

        self.transform = transforms.Compose(
            [
                transforms.Resize(
                    (image_size, image_size), interpolation=InterpolationMode.BICUBIC
                ),
                transforms.ToTensor(),
                self.normalize,
            ]
        )
\end{lstlisting}

The second class \texttt{BlipImageEvalProcessor} is designed to preprocess images for evaluation purposes. It resizes images to a specified size using bicubic interpolation, converts them to tensor format, and then normalizes them using the same mean and standard deviation values as the \texttt{BlipImageTrainProcessor}. 

\subsection{Implementation Details}
\label{implemnt details}
We selected the ViT-G/14 from EVA-CLIP~\citep{sun2023evaclip} as our visual encoder. The pre-trained weights are initialized and remain frozen during training. We removed the last layer from ViT~\citep{vit} and opted to use the output features of the second last layer, which yielded slightly better performance.
We first pre-train our patch embeddings projection layer using LLaVA filtered 558K image-text pairs from LAION~\citep{schuhmann2021laion400m}, CC-3M~\citep{sharma2018conceptual}, and SBU~\citep{sbu}, captioned by BLIP~\citep{li2022blip}. Using the pre-training stage leads to slightly better performance. During the vision-language instruction tuning stage, we initialize the Q-Former from InstructBLIP's weight and finetune the parameters of the Q-former and projection layer together while keeping both the image encoder and LLM frozen. We pre-trained the projection layer with 3 epochs with a batch size of 64. During the instruction finetuning stage, we employ a batch size of 24 with a maximum of 200K steps which roughly iterates two epochs of the training data. For both stage training, we used the AdamW~\citep{2017Decoupled} optimizer, with $\displaystyle \beta_1 = 0.9, \beta_2 = 0.999$, and a weight decay of 0.05. Additionally, we apply a linear warmup of the learning rate during the initial 1K steps, increasing from $\displaystyle 10^{-8}$ to $\displaystyle 10^{-5}$, followed by a cosine decay with a minimum learning rate of 0. The pre-training stage takes 6 hours and the instruction finetuning stage finished within two days on 8 Nvidia A6000 Ada (48G) GPUs. 

%\newpage also not allowed 

\subsection{Qualitative Examples}
\label{more examples}
% Additional Comparison between our model's performance and our baseline InstructBLIP. 

\begin{figure}[ht!]
\centering
\includegraphics[scale=0.7, width=\columnwidth]{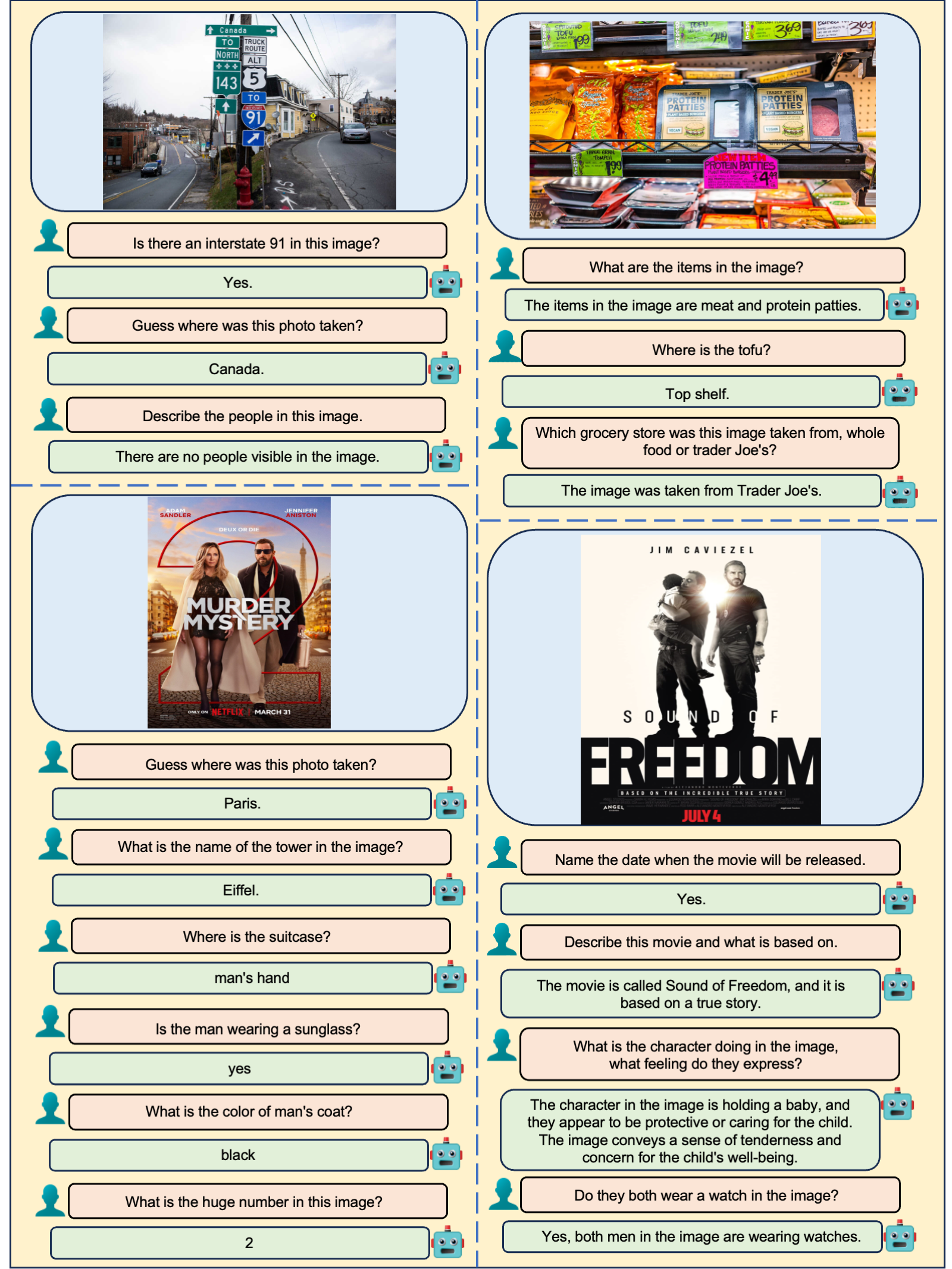}
\caption{Example of BLIVA's performance on real-life-scene and movie poster images. BLIVA's reply is strictly based on visual information with the ability to localize the objects in the image. BLIVA also demonstrates great OCR capabilities in reading road sign, food packaging, movie poster titles, and detailed texts.}
\label{fig:example-1}
\end{figure}

\begin{figure}[ht!]
\centering
\includegraphics[scale=0.9,  width=\columnwidth]{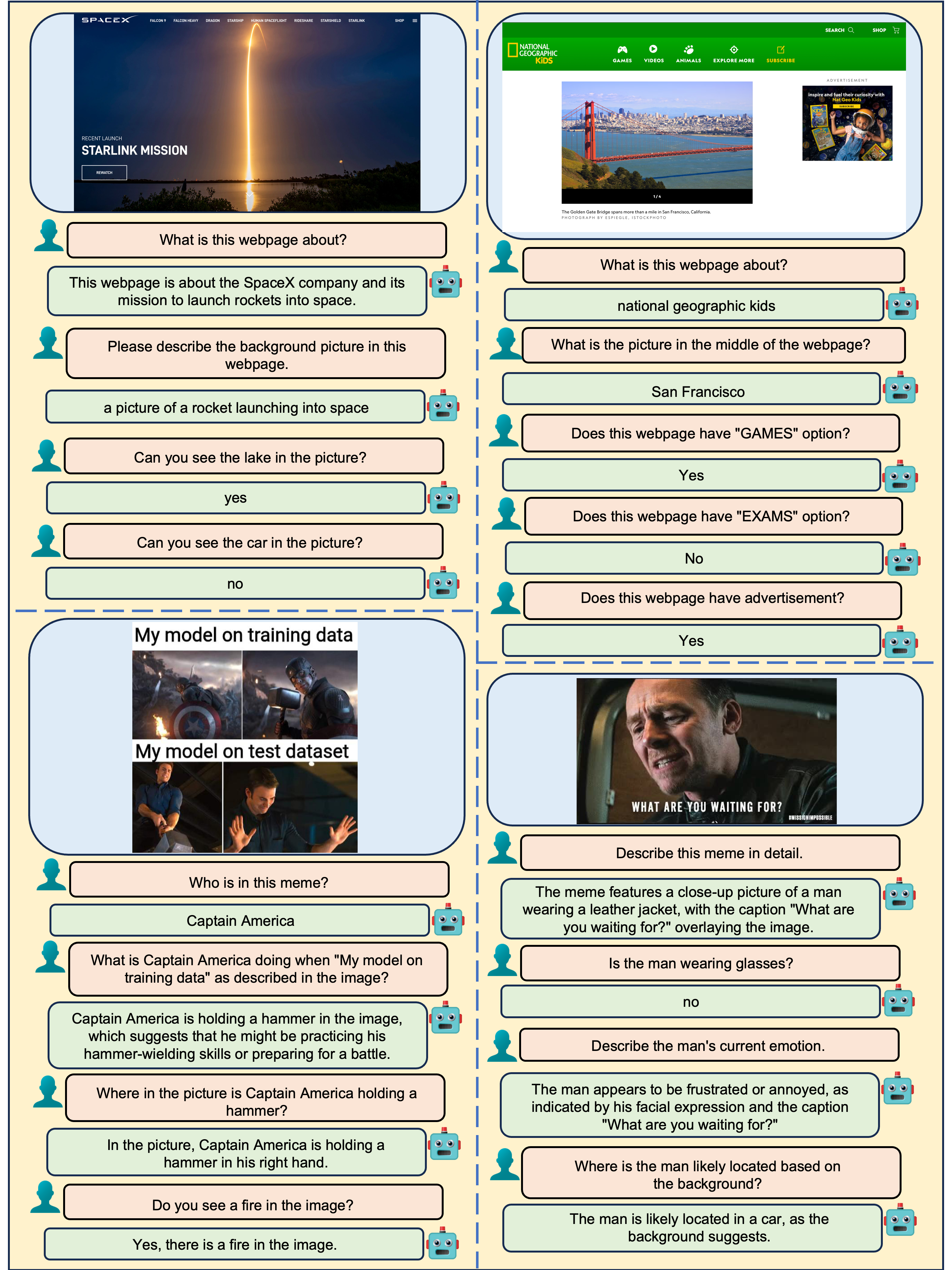}
\caption{Example of BLIVA's performance on the web page and meme images. BLIVA's reply shows its understanding of the visual information and the meaning behind both text and image. It can localize both the text and objects in the image clearly. BLIVA demonstrates great OCR capabilities in reading the text in memes and understanding the tabs on the web pages.}
\label{fig:example-2}
\end{figure}

Figure \ref{fig:example-1} and Figure \ref{fig:example-2} showcase the additional examples of our model's performance on various types of images, including the recognition of road signs and shopping products. Besides recognizing textual information, our models can also identify spatial relationships. These examples demonstrate the model's practical applicability and relevance in everyday life, as it can accurately interpret and analyze visual information that people commonly encounter, such as navigating traffic or identifying products in a store.

\subsection{YTTB-VQA Category distribution}

\begin{figure}[H]
\begin{center}
\includegraphics[scale=0.7, width=\columnwidth]{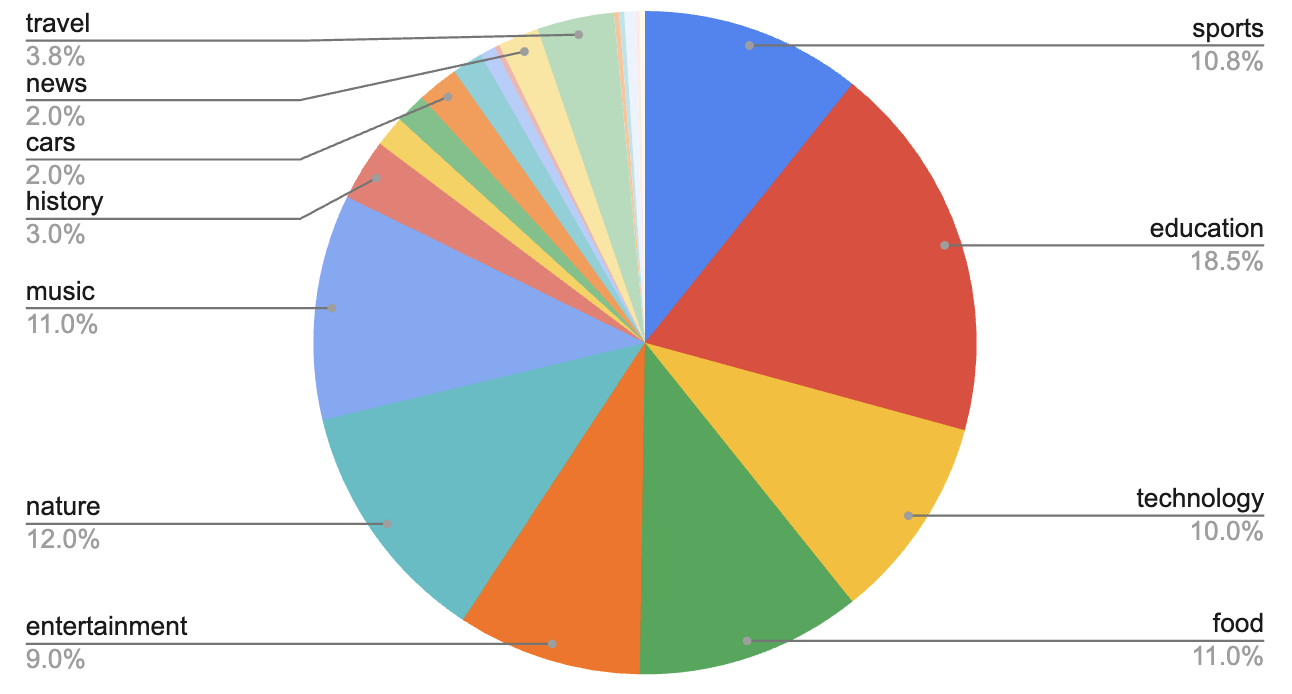}
\end{center}
\caption{\textbf{Category Distribution in YTTB-VQA}: This chart represents the distribution across 11 distinct categories within the YTTB-VQA dataset. These categories encompass a broad spectrum, including technology, sports, entertainment, food, news, history, music, nature, cars, and education.}
\label{sec:YTTB chart}
\end{figure}

\end{document}